\documentclass[sigconf]{acmart}

\newcommand{\resultone}[1]{\colorbox{green!15}{#1}}
\newcommand{\resulttwo}[1]{\colorbox{cyan!15}{#1}}
\newcommand{\resultthird}[1]{\colorbox{yellow!15}{#1}}
\newcommand{\bench}{\textsc{Pet-Bench}}

\usepackage{amssymb}
\usepackage{multirow}
\usepackage{algorithm}
\usepackage{algorithmic}
\usepackage{url}       
\AtBeginDocument{%
  }

\copyrightyear{2025}
\acmYear{2025}
\setcopyright{acmlicensed}\acmConference[CIKM '25]{Proceedings of the 34th ACM International Conference on Information and Knowledge Management}{November 10--14, 2025}{Seoul, Republic of Korea}

\acmBooktitle{Proceedings of the 34th ACM International Conference on Information and Knowledge Management (CIKM '25), November 10--14, 2025, Seoul, Republic of Korea}

\acmDOI{10.1145/3746252.3761605}
\acmISBN{979-8-4007-2040-6/2025/11}

\begin{document}

\title{Pet-Bench: Benchmarking the Abilities of Large Language Models as E-Pets in Social Network Services}

\author{Hongcheng Guo}
\authornote{Both authors contributed equally to this work.}
\affiliation{%
\institution{Fudan University}
  \institution{School of Data Science}
  \city{Shanghai}
  \country{China}
}
\email{guohc@fudan.edu.cn}

\author{Zheyong Xie}
\authornotemark[1]
\affiliation{%
  \institution{Xiaohongshu Inc.}
  \department{NLP Team}
  \city{Beijing}
  \country{China}
}
\email{xiezheyong@xiaohongshu.com}

\author{Shaosheng Cao}
\affiliation{%
  \institution{Xiaohongshu Inc.}
  \department{NLP Team}
  \city{Beijing}
  \country{China}
}

\email{caoshaosheng@xiaohongshu.com}
\authornote{Corresponding author.}

\author{Boyang Wang}
\affiliation{%
  \institution{Beihang University}
  \department{CCSE}
  \city{Beijing}
  \country{China}
}
\email{wangboyang@buaa.edu.cn}

\author{Weiting Liu}

\affiliation{%
  \institution{Fudan University}
\department{Institute of Science and Technology for Brain-inspired Intelligence}
\city{Shanghai}
\country{China}
}
\email{weitinglau1999@gmail.com}

\author{Zheyu Ye}
\affiliation{%
  \institution{Xiaohongshu Inc.}
  \department{NLP Team}
  \city{Beijing}
  \country{China}
}
\email{zheyuye@xiaohongshu.com}

\author{Zhoujun Li}
\affiliation{%
  \institution{Beihang University}
  \department{CCSE}
  \city{Beijing}
  \country{China}
}
\email{lizj@buaa.edu.cn}

\author{Zuozhu Liu}

\affiliation{%
  \institution{Zhejiang University}
  \department{ZJU-UIUC Institute}
    \city{Haining}
  \country{China}
}
\email{zuozhuliu@intl.zju.edu.cn}

\author{Wei Lu}

\affiliation{%
  \institution{Singapore University of Technology and Design}
  \department{StatNLP Research Group}
  \city{Singapore}
  \country{Singapore}
}
\email{luwei@sutd.edu.sg}

\renewcommand{\shortauthors}{Hongcheng Guo et al.}

\begin{abstract}
As interest in using Large Language Models for interactive and emotionally rich experiences grows, virtual pet companionship emerges as a novel yet underexplored application. Existing approaches focus on basic pet role-playing interactions without systematically benchmarking LLMs for comprehensive companionship. In this paper, we introduce \bench{}, a dedicated benchmark that evaluates LLMs across both \textbf{\textit{self-interaction}} and \textbf{\textit{human-interaction}} dimensions. Unlike prior work, \bench{} emphasizes self-evolution and developmental behaviors alongside interactive engagement, offering a more realistic reflection of pet companionship. It features diverse tasks such as intelligent scheduling, memory-based dialogues, and psychological conversations, with over 7,500 interaction instances designed to simulate pet behaviors. Evaluation of 28 LLMs reveals significant performance variations linked to model size and inherent capabilities, underscoring the need for specialized optimization in this domain. \bench{} serves as a foundational resource for benchmarking pet-related LLM abilities and advancing emotionally immersive human-pet interactions.~\footnote{Code and dataset are available at \url{https://github.com/HC-Guo/Act-as-Pet}.}

\end{abstract}

\begin{CCSXML}
<ccs2012>
   <concept>
       <concept_id>10010147.10010178.10010179</concept_id>
       <concept_desc>Computing methodologies~Natural language processing</concept_desc>
       <concept_significance>500</concept_significance>
       </concept>
 </ccs2012>
\end{CCSXML}

\ccsdesc[500]{Computing methodologies~Natural language processing}

\keywords{Benchmark, Large Language Model, Emotional Support, E-Pet}
  
\maketitle

\section{Introduction}

\begin{figure*}
    \centering
    \includegraphics[width=0.65\linewidth]{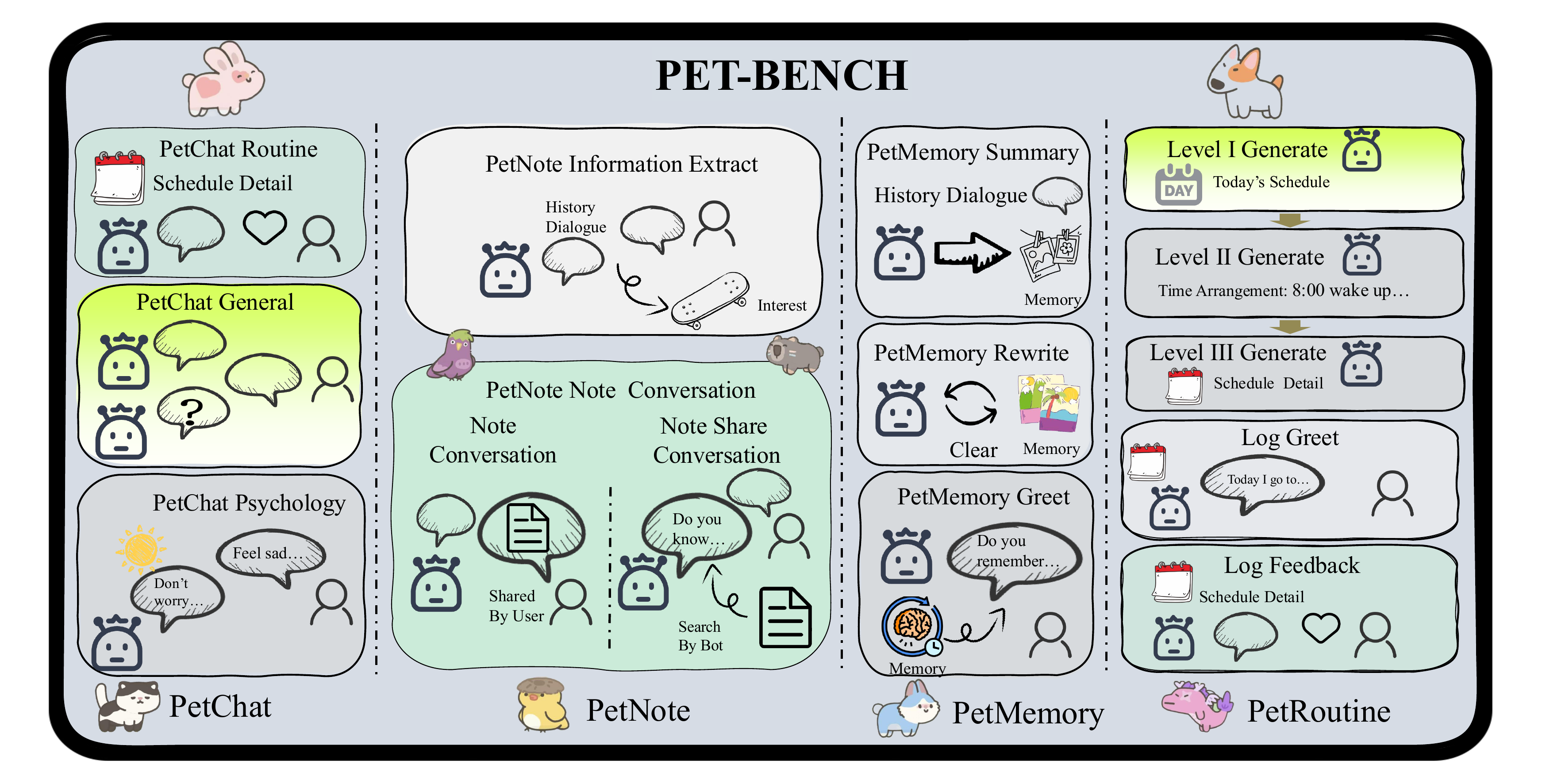}
    % \vspace{-1.1em}
    
    \label{fig:overview}
    \caption{Overview of \bench{}. The framework consists of four key components: PetChat, PetNote, PetMemory, and PetRoutine. }
\end{figure*}

With the increasing interest in utilizing Large Language Models (LLMs) for interactive and emotionally enriched experiences~\cite{wang2023rolellm,shao2023character,qiu2024interactive}, the application of LLMs to pet companionship scenarios on social network services (SNS) has attracted significant attention. However, current practices~\cite{xu2024can, treder2024introduction, working-memory-2024, maharana2024evaluating, zhong2024memorybank} predominantly rely on simple, role-playing dialogues where LLMs directly mimic pet behaviors without internal reflection or adaptation~\cite{wang2024incharacter,xu2024character}. Such modes result in static interaction patterns, lacking long-term memory, developmental progression, and adaptive emotional responses crucial to authentic companion. Thus, this area requires a systematic benchmark to better assess the advancement of LLM capabilities.

To bridge these gaps, we introduce \bench{}, a pioneering benchmark specifically designed to assess performance of LLMs as E-pet on social network services (SNS). Unlike prior work confined to momentary alignment between persona and actions~\cite{wang2023rolellm,shao2023character,chen2024persona}, \bench{} emphasizes self-interaction, highlighting the importance of internal reflection, self-evolution, and adaptive emotional behaviors in addition to interactive engagement. This approach mirrors authentic virtual pet companionship, where pets continuously evolve through interactions, adapt to emotions of users, and gradually develop unique traits over time.

\bench{} features four key tasks: \textbf{\textit{PetChat}}: Facilitates general conversations with both pets and humans, including psychological interactions and dialogues based on the routine of pets. 
\textbf{\textit{PetNote}}: Focuses on tasks related to social media notes, such as information extraction, intent recognition, retrieval-augmented dialogue.
\textbf{\textit{PetMemory}}: Handles memory-related tasks such as summarization, rewriting, and initiating human interactions based on stored experiences.
\textbf{\textit{PetRoutine}}: Simulates the daily schedule of pets at three levels of detail, with higher levels providing more granular plans.

\bench{} comprises over 7,500 instances designed to simulate lifelike pet interactions. We conduct experiments on \bench{} using more than 20 prominent LLMs, leading to the following main contributions:

\begin{itemize}
    \item We propose \textbf{\bench{}}, the first benchmark designed to evaluate LLM-driven pet companionship capabilities across multiple nuanced tasks on social network services.
    \item The comprehensive evaluation is conducted spanning different sizes and series. Generally, closed-source models outperform open-source ones. Nevertheless, the performance gap between the leading closed-source model and the top open-source model is quite small, only about 1\%.
    \item Our findings show that models perform more poorly in tasks that involve understanding complex emotions or long memories, compared with more straightforward tasks.
\end{itemize}

\section{Related Work}
\paragraph{Role-Playing with LLM}
LLMs have exhibited remarkable language understanding and generation proficiency, catalyzing significant role-playing application advancements~\cite{tseng2024two, chen2024persona}. With the continuous evolution of research paths, the academic community has shifted towards utilizing and optimizing large language models to accurately reproduce the multidimensional characteristics of roles. Domain-specific knowledge repositories \citep{li2023chatharuhi,chen2023large,wang2023rolellm}, distinctive linguistic expression patterns \citep{wang2023rolellm,zhou2023characterglm}, sophisticated cognitive decision frameworks \citep{zhao2023narrativeplay,xu2024character}, and nuanced personality dimensions \citep{shao2023character,wang2024incharacter} are all encompassed within these.

\paragraph{From Traditional to LLM Pet Companion}
The human-pet bond enhances mental well-being by reducing loneliness and stress \cite{animals, ani14030441, FLYNN2020101223, AbatRoy2021}, yet constraints like space, allergies, and pet abandonment \cite{Costa2022, Horecka2022} drive demand for alternatives. While robotic \cite{Coghlan2021, Hwang2024, Jeong2023} and virtual pets \cite{Zhou2024} exist, their rule-based designs \cite{Baudier2023, Bengani2024} lack emotional depth, and early NLP chatbots struggle with semantic understanding and connection \cite{Rasool2024, Wei2024}.  
LLMs enable context-aware, personalized interactions \cite{yu2024experimental, guo2024large, stade2024large, qiu2024interactive}, with multimodal integration enhancing realism \cite{xie2024large}. However, LLM-based virtual pets \cite{xu2024can, treder2024introduction} still face persona inconsistency \cite{sun2024building}, latency, multimodal challenges, limited episodic memory \cite{working-memory-2024, maharana2024evaluating, zhong2024memorybank}, and ethical concerns \cite{ethical2024current, jiao2024navigating}. 
We address these gaps with 
% a novel self-interaction mechanism, human-in-the-loop feedback, and 
a comprehensive evaluation framework supported by an expanded dataset.

\begin{figure*}
    \centering
    \includegraphics[width=0.65\textwidth]{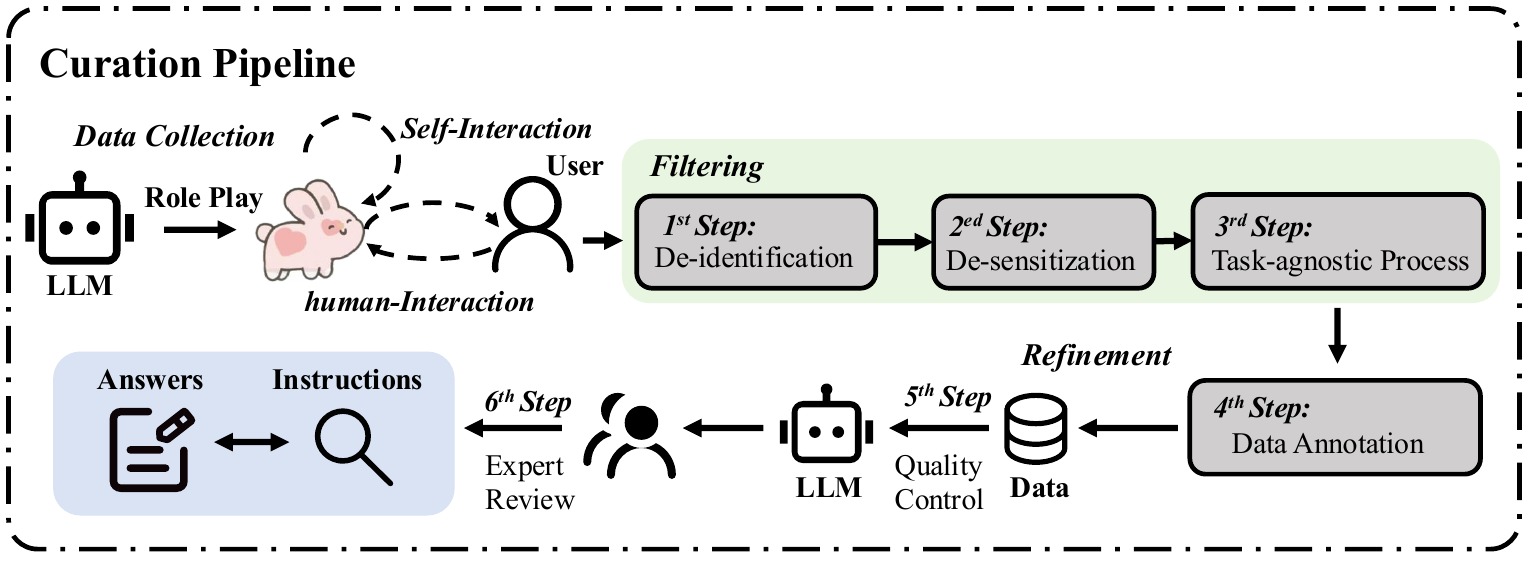}
    \caption{Pipeline for data curation in \bench{}, organized into three phases.}
    \label{fig:pipline}
\end{figure*}

\vspace{-0.5em}
\section{\bench{}}

In Section~\ref{sec:define}, we present a comprehensive overview of E-Pet’s core capabilities as shown in Figure \ref{fig:overview}, while Section~\ref{sec:build} elaborates on the systematic pipeline developed for curating benchmark data.
\vspace{-0.5em}

\subsection{Definition} \label{sec:define}

\bench{} is designed around two fundamental interaction modes for LLM-driven pet companionship: \textbf{\textit{human-interaction}} and \textbf{\textit{self-interaction}}. Human-interaction enables meaningful engagement with users through dialogue and responsive behaviors, while self-interaction allows the model to autonomously simulate pet-like behaviors, routines, and memories. These modes are systematically organized into four core dimensions: \textbf{PetChat}, \textbf{PetNote}, \textbf{PetMemory}, and \textbf{PetRoutine}, where PetChat and PetNote focus on interactions with the user, and PetMemory and PetRoutine pertain to the model’s autonomous internal processes.
    
 $\bullet$ \textbf{PetChat.} This capability assesses conversational interactions between virtual pets and humans, including: 
 1) General Dialogue -- Basic interactions and conversations initiated by humans. 
 2) Psychological Dialogue -- Understanding and responding to emotional cues during interactions. 
 3) Contextual Routine Dialogue -- Engaging in conversations informed by previous daily routines.

 $\bullet$ \textbf{PetNote.} Understanding and engaging with pet-related social media content involves:
 4) Information Extraction -- Deriving essential details from social media notes. 
 5) Note-share Conversation -- Engaging in discussions according to user-shared notes.
 6) Intent Recognition -- Determining whether external retrieval is needed based on user intent.
 7) Note-based Conversation -- Responding to user questions according to  note content.
 8) Note-based Greeting -- Initiating interactions inspired by pet-relevant social media notes.

 $\bullet$ \textbf{PetMemory.} Effective pet companionship involves memory-driven interactions, including: 9) Memory Summarization -- Summarizing interactions or events. 10) Memory Rewriting -- Reconstructing pet experiences into coherent narratives. 
 11) Memory-based Dialogue -- Proactively engaging with humans based on remembered experiences.

 $\bullet$ \textbf{PetRoutine.} Creating engaging virtual pets necessitates realistic routine implementation, including: 12) Routine Level I/II/III Generation -- Generating diverse pet routines across three different complexity levels. 13) Routine-based Greet -- Initiating interactions with humans based on routines. 14) Routine-based Feedback -- Generating responses based on users’ perspectives on a pet’s routines.

\subsection{Construction} \label{sec:build}
In this section, we outline the process of constructing \bench{}, which comprises three main phases: (1) Data Collection, (2) Filtering, and (3) Refinement, as shown in Figure \ref{fig:pipline}.

\subsubsection{Collection}

To evaluate large language models in pet companionship simulation, we use data generated from detailed simulations reflecting realistic pet-owner interactions.

We collect over 7,500 interaction dialogues and routines covering pet-owner conversations, emotional expressions, routine activities, memory summaries, and intent annotations. 
To ensure diversity in our evaluation process, we incorporate multiple dimensions: 
\textbf{(1) Interaction Diversity.} Scenarios encompass dialogues, emotional support, routine care (feeding, exercise, grooming), and memory-based tasks. 
\textbf{(2) Pet Type Diversity.} Simulated species include dogs, cats, rabbits, birds, and reptiles, capturing a range of behavioral and emotional characteristics. 
\textbf{(3) Complexity Levels.} Tasks range from basic routines to complex adaptive behaviors, enabling evaluation across varying difficulty levels. 
Based on these considerations, we systematically collect comprehensive data for further processing.

\subsubsection{Filtering}

After data collection, the dataset undergoes preliminary filtering to ensure quality: (1) User identifiers and personal information are removed to protect privacy; (2) Politically sensitive, inappropriate, or offensive content is excluded; (3) Irrelevant or low-quality simulation data is eliminated using task-agnostic rules to maintain relevance and reliability. 

\begin{table}[h!]
    \centering
    \resizebox{\linewidth}{!}{%
    \begin{tabular}{@{}lc|lc@{}}
        \toprule
        \textbf{Statistics} & \textbf{Value} & \textbf{Statistics} & \textbf{Value} \\ \midrule
        PetChat General & 2,511    & PetMemory Summary  & 412 \\
        PetChat Psychology & 100   & PetMemory Rewrite & 589 \\
        PetChat Routine & 533      & PetMemory Greet & 102 \\
        PetNote Information Extract & 511 & PetRoutine Level I & 362   \\
        PetNote Share Conversation & 475 & PetRoutine Level II & 444 \\
        PetNote Intent Recognition & 1,114 & PetRoutine Level III & 143 \\ 
        PetNote Conversation & 192 & PetRoutine Feedback  & 163 \\
        PetNote Greet & 88 & PetRoutine Greet     & 76  \\
        \midrule
        \textbf{Total Questions}                 & \textbf{7,815} & 
        \textbf{Total Input Tokens}              & \textbf{4,288,651} \\
        \textbf{Avg. Input length (words)} & \textbf{548.77}  & 
        \textbf{Total Output Tokens}             & \textbf{192,533} \\
        \textbf{Avg. Output length (words)}      & \textbf{24.64} & & \\
        \bottomrule
    \end{tabular}
    }
    \vspace{1em}
    \caption{Statistics of \bench{} Dataset. }
    \label{tab:quantity}
\end{table}
\vspace{-3em}

% \vspace{-2em}
\subsubsection{Refinement}
To obtain higher-quality and more reliable data, we incorporated multiple refinement steps, as detailed below:
\textbf{(1) Data Annotation.} During the data annotation phase, every data entry is carefully checked and refined by human annotators to ensure consistency with our evaluation goals.
\textbf{(2) Quality Control.}
To ensure high data quality, we introduce an additional LLM-based quality evaluation step. Specifically, all items are scored using GPT-4~\citep{gpt4}, and instances identified as low quality are then reviewed manually to enable efficient filtering and quality control. 
\textbf{(3) Expert Review.} 
As a final validation step, the dataset is rigorously reviewed by 10 expert annotators, with each entry independently assessed by at least three reviewers for coherence and domain relevance. Disagreements are resolved by majority vote.

By following this comprehensive pipeline, we have constructed an extensive and meticulously validated dataset for rigorous ability evaluation, with statistics summarized in Table~\ref{tab:quantity}.

% --v1.0-

\begin{table*}[htbp]
\centering
\resizebox{1\textwidth}{!}{%
\begin{tabular}{cccc|ccccc|ccc|ccccc|c}
\toprule
\multirow{2}{*}{\textbf{Models}} &
  \multicolumn{3}{c}{\textbf{PetChat}} &
  \multicolumn{5}{c}{\textbf{PetRoutine}} &
  \multicolumn{3}{c}{\textbf{PetMemory}} &
  \multicolumn{5}{c}{\textbf{PetNote}} &
  \textbf{Avg.} \\ 
  \cline{2-18}\addlinespace[2pt] 
 &
  \textbf{General} &
  \textbf{Psychological} &
  \textbf{Routine} &
  \textbf{Level I} &
  \textbf{Level II} &
  \textbf{Level III} &
  \textbf{Feedback} &
  \textbf{Greet} &
  \textbf{Summary} &
  \textbf{Rewrite} &
  \textbf{Greet} &
  \textbf{Information Extract} &
  \textbf{Note Conversation} &
  \textbf{Intent Recognition} &
  \textbf{Retrieve Conversation} &
  \textbf{Greet} &
  \textbf{-} \\ \midrule
        % ==========================================
        \multicolumn{18}{c}{\textit{Open-Source Large Language Models (1.5B+)}} \\ \midrule

         Qwen-2.5-1.5B & 11.73 & 15.46 & 13.62 & 54.69 & 65.81 & 60.58 & 12.20 & 12.15 & 15.79 & 28.21 & 14.70 & 21.63 & 15.30 & 18.46 & 13.76 & 16.98 & 24.44 \\

        Qwen-2.5-1.5B-Instruct & 13.45 & 17.37 & 15.80 & 63.25 & 73.59 & 66.02 & 14.78 & 17.51 & 19.91 & 39.20 & 17.37 & 48.85 & 32.79 & 40.82 & 13.19 & 20.16 & \resulttwo{32.13} \\

         Llama-3.2-1B & 8.70 & 8.80 & 10.19 & 49.09 & 55.68 & 46.98 & 6.35 & 6.60 & 10.30 & 11.06 & 6.62 & 7.49 & 7.73 & 7.61 & 8.01 & 8.94 & 16.26 \\

         Llama-3.2-1B-Instruct & 13.84 & 15.93 & 16.31 & 57.13 & 61.77 & 62.84 & 13.79 & 15.17 & 14.46 & 21.42 & 16.50 & 23.96 & 12.55 & 18.26 & 14.80 & 21.15 & \resultthird{24.99} \\

         Phi-3.5-Mini-Instruct-3.82B & 16.19 & 19.16 & 18.80 & 72.13 & 75.99 & 78.18 & 14.33 & 18.61 & 14.54 & 33.33 & 19.38 & 48.40 & 22.23 & 35.32 & 18.98 & 21.25 & \resultone{32.93} \\

        % ==========================================
        \midrule
        \multicolumn{18}{c}{\textit{Open-Source Large Language Models (7B+)}} \\ \midrule

        Qwen-2.5-7B & 12.52 & 19.65 & 14.73 & 61.66 & 75.80 & 78.99 & 14.00 & 16.17 & 19.56 & 29.24 & 16.62 & 34.84 & 18.92 & 26.88 & 15.59 & 22.04 & 29.83 \\

        Qwen-2.5-7B-Instruct & 15.05 & 19.82 & 18.12 & 67.14 & 76.48 & 85.14 & 14.83 & 19.49 & 42.90 & 39.29 & 18.92 & 49.94 & 45.71 & 47.83 & 17.28 & 23.57 & \resultone{37.59} \\ 

        Internlm-2.5-7b-Chat & 16.77 & 21.04 & 20.18 & 69.89 & 76.59 & 82.28 & 15.15 & 18.11 & 25.44 & 36.22 & 17.12 & 37.88 & 37.85 & 37.87 & 19.76 & 20.62 & 34.55 \\

        Llama-3.1-8B & 8.60 & 7.14 & 10.86 & 47.43 & 55.64 & 45.50 & 6.44 & 6.58 & 8.35 & 7.84 & 6.91 & 10.15 & 10.06 & 10.11 & 10.86 & 10.40 & 16.43 \\

        Llama-3.1-8B-Instruct & 16.91 & 20.11 & 18.39 & 67.59 & 71.93 & 84.70 & 16.07 & 19.69 & 21.28 & 39.96 & 20.07 & 48.45 & 40.44 & 44.44 & 17.14 & 24.02 & 35.70 \\

        Internlm-3-8b-Instruct & 16.96 & 20.62 & 18.87 & 63.57 & 76.42 & 84.71 & 15.16 & 18.74 & 37.93 & 33.64 & 16.98 & 50.91 & 43.01 & 46.96 & 18.40 & 22.51 & 36.59 \\

        Glm-4-9B-Chat & 18.17 & 20.27 & 20.47 & 66.61 & 77.78 & 86.04 & 15.60 & 19.40 & 23.28 & 31.59 & 17.77 & 53.77 & 42.64 & 48.21 & 20.50 & 24.15 & \resultthird{36.64} \\

        Phi-4-14B & 18.43 & \textbf{21.80} & 20.94 & 69.39 & 79.68 & 87.14 & 16.17 & 20.10 & 38.01 & 32.09 & 18.96 & 48.58 & 36.06 & 42.32 & \textbf{22.75} & 23.58 & \resulttwo{37.25} \\

        Internlm-2.5-20b-Chat & 16.67 & 20.59 & 18.52 & 70.30 & 78.61 & 86.44 & 14.88 & 18.68 & 21.49 & 35.09 & 15.67 & 47.30 & 38.54 & 42.92 & 19.77 & 22.29 & 35.48 \\

        Gemma-2-27B-It & 12.06 & 14.99 & 13.22 & 47.53 & 50.44 & 46.36 & 10.57 & 10.60 & 10.95 & 11.74 & 10.79 & 13.80 & 9.64 & 11.72 & 13.91 & 14.41 & 18.92 \\

        % ==========================================
        \midrule
        \multicolumn{18}{c}{\textit{Open-Source Large Language Models (32B+)}} \\ \midrule

         Qwen-2.5-32B & 14.92 & 21.07 & 15.81 & 60.77 & 75.93 & 79.40 & 14.64 & 18.97 & 19.11 & 28.30 & 18.76 & 42.84 & 31.67 & 37.25 & 17.25 & 21.58 & 32.39 \\

         Qwen-2.5-32B-Instruct& 18.21 & 20.88 & 21.25 & 68.93 & 77.42 & 87.02 & 16.19 & 19.51 & 50.39 & 39.32 & 19.21 & 53.54 & 47.11 & 50.32 & 19.17 & 23.59 & \resultthird{39.50} \\

         Yi-1.5-34B-Chat & 17.28 & 20.61 & 19.30 & 71.77 & 76.52 & 85.71 & 16.34 & 19.91 & 38.17 & 32.73 & 19.36 & 52.37 & 44.23 & 48.30 & 19.82 & 22.77 & 37.82 \\

         Llama-3.3-70B-Instruct & 14.75 & 15.38 & 13.81 & 60.24 & 76.25 & 85.43 & 16.24 & \textbf{20.42} & 41.38 & 35.50 & \textbf{20.16} & 51.12 & 50.75 & 50.93 & 13.34 & 20.99 & 36.67 \\

         Qwen-2.5-72B & 14.72 & 20.87 & 15.70 & 58.94 & 75.66 & 81.34 & 16.10 & 18.69 & 20.76 & 27.28 & 19.06 & 39.73 & 20.24 & 29.98 & 14.40 & 21.82 & 30.96 \\

         Qwen-2.5-72B-Instruct & \textbf{19.28} & 20.94 & \textbf{21.74} & 69.96 & 79.48 & 88.72 & \textbf{17.14} & 19.78 & \textbf{51.31} & 42.75 & 19.22 & 56.66 & 50.19 & 53.42 & 19.69 & 23.72 & \resultone{40.87} \\

         Deepseek-V3& 18.37 & 21.11 & 20.85 & 67.27 & 79.12 & 86.14 & 16.09 & 19.81 & 49.34 & 39.89 & 17.96 & 56.69 & 47.43 & 52.06 & 18.90 & \textbf{24.54} & \resulttwo{39.72} \\

         Deepseek-R1 & 17.22 & 17.30 & 19.41 & \textbf{74.47} & 68.43 & 88.40 & 14.30 & 17.75 & 36.46 & 28.45 & 18.34 & 55.34 & 47.26 & 51.30 & 18.83 & 19.77 & 37.06 \\ 

        % ==========================================
         \midrule
        \multicolumn{18}{c}{\textit{Closed-Source Large Language Models (API)}} \\ \midrule

        GPT-4o& 18.69 & 20.21 & 20.70 & 71.97 & \textbf{79.69} & 89.67 & 16.00 & 18.89 & 48.92 & \textbf{44.56} & 17.95 & \textbf{56.78} & \textbf{55.24} & \textbf{56.01} & 20.65 & 23.28 & \resultone{41.20} \\

        GPT-4o-Mini & 18.48 & 19.27 & 20.69 & 68.82 & 78.63 & 89.90 & 16.97 & 19.75 & 49.15 & 34.45 & 19.13 & 54.68 & 41.22 & 47.95 & 20.77 & 24.02 &  \resultthird{38.99} \\

        Glm-4-Plus & 18.55 & 20.35 & 20.51 & 70.46 & 79.17 & 89.32 & 17.08 & 19.40 & 50.80 & 38.10 & 19.06 & 54.05 & 47.48 & 50.76 & 18.77 & 23.63 &  \resulttwo{39.84} \\

        Claude-3.5-Sonnet & 19.03 & 19.09 & 21.60 & 67.60 & 79.28 & \textbf{90.10} & 15.62 & 18.77 & 18.23 & 32.32 & 17.58 & 52.54 & 48.16 & 50.35 & 19.63 & 19.86 & 36.86  \\

        Gemini-1.5-Pro-002 & 16.86 & 16.05 & 17.84 & 61.99 & 78.67 & 89.37 & 16.61 & 20.11 & 41.75 & 32.11 & 19.76 & 55.90 & 49.62 & 52.76 & 17.72 & 20.46 & 37.97 \\

\bottomrule
\end{tabular}%
}
\caption{Results of different models on the 
\bench{}. We utilize \resultone{green}(1st) \resulttwo{blue}(2nd) \resultthird{yellow}(3rd) to distinguish the top three results within different sizes.}
\label{tab:my-table}
\end{table*}

\section{Experiments} \label{sec:experiment}

\subsection{Evaluation Settings}

Experiments are carried out using 128 NVIDIA H800 GPUs, where 28 models, comprising both open-source and closed-source models, are tested on the \bench{} benchmark, as illustrated in Table \ref{tab:my-table}.

\subsection{Evaluation Protocol}

For all tasks except PetNote Intent Recognition and PetRoutine Level I/II/III, The metric is the average of seven metrics:  
\begin{equation}
S = \frac{1}{7} \left( \sum_{n=1}^{4} \scriptsize{\text{BLEU-}}n + \scriptsize{\text{ROUGE-1}} + \scriptsize{\text{ROUGE-L}} + \scriptsize{\text{S}} \right)
\end{equation}
where BLEU-n~\cite{papineni-etal-2002-bleu} and ROUGE~\cite{lin2004rouge} are common metrics for generation evaluation, and $S$ denotes the semantic similarity between the predicted and target sentences, calculated using BGE~\cite{bge_m3}.

For the PetNote Intent Recognition task, accuracy is included among the evaluation metrics, while for PetRoutine Levels I, II, and III, GPT-4o scores are used as supplementary evaluation measures.

\subsection{Main Results}
\subsubsection{Overall Evaluation}
The benchmark results highlight key insights into LLM performance on diverse tasks.  
\textbf{Closed-Source Models Excel:} Proprietary models consistently outperform open-source counterparts, benefiting from superior training resources and optimization techniques. \textbf{\textit{GPT-4o}} leads with an average score of \textbf{41.20}, closely followed by \textbf{\textit{Glm-4-Plus}} (39.84).
\textbf{Task Complexity Varies:} Tasks like \textit{PetChat-General } and \textit{PetMemory-Summary} show strong results across models. In contrast, tasks like \textit{PetMemory-Rewrite} pose significant challenges, with even top models struggling.
\textbf{Scale Enhances Performance:} Among open-source models, larger architectures like \textbf{\textit{Qwen-2.5-72B-Instruct}} achieve the best average score of \textbf{40.87}, underscoring the importance of model size in tackling diverse \bench{} tasks, as shown in Table \ref{tab:my-table}. 

\subsubsection{Task-Specific Evaluation}
Multi-task evaluation reveals three critical insights about LLM performance.
 \textbf{(1) Specialized Task Dominance.} Current models exhibit strong task specialization but lack cross-scenario generalization. \textit{Phi-4-14B} and \textit{Deepseek-V3} dominate structured tasks, while \textit{Qwen-72B-Instruct} excels in contextual interactions.
 \textbf{(2) Complexity-Driven Performance Gap.} 
 Complex interaction tasks in PetChat achieve low average accuracy, whereas structured tasks like routine generation maintain high accuracy levels. 
 \textbf{(3) Stability-Specificity Trade-off.} \textit{Deepseek-V3} and \textit{Phi-4-14B} demonstrate robust stability, particularly in memory-related tasks. \textit{Qwen} variants show polarized performance through instruction tuning, yet limited memory rewriting improvements. \textit{GPT-4o} emerges as the most balanced closed-source model.

\section{Conclusion}

We introduce \bench{}, a benchmark for evaluating LLMs in pet companionship across \textbf{\textit{self-interaction}} and \textbf{\textit{human-interaction}}. \bench{} emphasizes self-evolution and developmental behaviors, offering a more comprehensive assessment. With diverse tasks and over 7,500 instances, our evaluation of 28 LLMs reveals performance gaps, highlighting the need for further optimization. 

\section*{Acknowledgments}
We would like to thank the anonymous reviewers for their constructive comments on our work. This research/project is partially supported by the National Research Foundation, Singapore under its National Large Language Models Funding Initiative. (AISG Award No: AISG-NMLP-2024-005).

\clearpage

\section*{GenAI Usage Disclosure}
As detailed in Section \ref{sec:build}, generative AI was employed as an essential tool in the data building phase. During the writing process, generative AI was also used to enhance the quality of the manuscript, including correcting spelling and grammatical errors.

\bibliographystyle{ACM-Reference-Format}
% \balance
\bibliography{custom}

% \appendix

% \input{contents/appendix}

\end{document}